# LIFE-INSPIRED INTEROCEPTIVE ARTIFICIAL INTELLIGENCE

## FOR AUTONOMOUS AND ADAPTIVE AGENTS

Sungwoo Lee[1,2,3,4*], Younghyun Oh[1,2,3,4*], Hyunhoe An[1,2,3,4], Hyebhin Yoon[1,2,3,4],

Karl J. Friston[5], Seok Jun Hong[1,2,3,4,6†], Choong-Wan Woo[1,2,3,4†]

[1] Center for Neuroscience Imaging Research, Institute for Basic Science, Suwon, South Korea
[2] Department of Biomedical Engineering, Sungkyunkwan University, Suwon, South Korea
[3] Department of Intelligent Precision Healthcare Convergence, Sungkyunkwan University, Suwon, South Korea
[4] Life-inspired Neural Network for Prediction and Optimization Research Group, Suwon, South Korea
[5] Wellcome Centre for Human Neuroimaging, Institute of Neurology, University College London, United Kingdom
[6] Center for the Developing Brain, Child Mind Institute, NY, USA

[*]co-first authors, [†]co-corresponding authors

**Running Head:** INTEROCEPTIVE AI

Please address correspondence to:
Choong-Wan Woo
Email: waniwoo@skku.edu
Telephone: +82 (31) 299-4363

Seok-Jun Hong
Email: hongseokjun@g.skku.edu
Telephone: +82 (31) 299-4345

Department of Biomedical Engineering
Center for Neuroscience Imaging Research
Sungkyunkwan University
Suwon 16419, Republic of Korea

## ABSTRACT

Building autonomous (i.e., choosing goals based on their needs) and adaptive (i.e., surviving in ever-changing environments) agents has been a long-standing challenge in artificial intelligence (AI). Living organisms exemplify such agents, offering valuable insights into adaptive autonomy. Here, we focus on interoception—the process of monitoring one's internal environment to keep it within certain bounds, which underwrites the survival of an organism. To develop AI with interoception, it is necessary to factorize state variables representing internal and external environments and incorporate life-inspired mathematical properties of internal states. Additionally, key functionalities related to interoception, such as neuromodulatory mechanisms, can be implemented to enable internal context-dependent adaptive behaviors. This paper presents a new perspective on how interoception can enhance autonomous and adaptive AI by integrating the legacy of cybernetics with recent advances in theories of life, reinforcement learning, and neuroscience.

*"Any autonomous agent needs to be able to resolve two things: what to do next and how to do it"*

by Spier & McFarland [1]

*"If an agent is to sustain itself over extended periods of time in a continuously changing, unpredictable environment, it must be adaptive"*

by Pfeifer and Scheier [2]

## LEARNING FROM LIVING ORGANISMS TO BUILD AUTONOMOUS AND ADAPTIVE INTELLIGENCE

Although significant advances in artificial intelligence (AI) have led to remarkable achievements in many domains [3-5], building autonomous and adaptive agents still requires further attention. To be autonomous, agents must make decisions and execute actions depending on their own goals. To be adaptive, agents must be able to dynamically reconfigure their state representations in response to changing environments. Conventional artificial agents, however, rely on pre-engineered inputs to establish new goals [6] and are often incapable of adapting to environmental changes [7]. These challenges are particularly important as many industrial fields increasingly demand more capable agents for complex tasks, for example, robot exploration and rescue tasks in the space. In these tasks, robots must manage system health and complete missions autonomously with minimal human intervention [8,9]. To overcome these challenges, new ideas from other fields need to be integrated into existing AI frameworks, and we can draw inspiration from living organisms. These organisms exhibit inherent autonomy and adaptivity—for instance, even single-cell eukaryotes can adjust their goals based on their needs and varying conditions [10]. Additionally, living organisms possess an innate drive to avoid or escape hostile environments to preserve their well-being. In essence, living agents inherently know how to choose their own goals and adapt their behaviors in response to changing environments, with the primary goal of maintaining homeostasis within their internal environment—a crucial factor for survival (**Fig. 1A**).

Here, we propose an interoceptive AI framework that endows an artificial agent with an internal environment and interoceptive inputs, which could provide contextual information that supports both autonomy and adaptivity. This framework aims to build an agent that can maintain homeostasis within its internal environment state, which requires the ability to monitor its internal state (i.e., interoception). In developing this framework, we integrate the original concepts from cybernetics and combine them with recent theoretical work on life (**Box 1**) and reinforcement learning (RL) [11,12] to formalize the internal state as a "universal and valuable context"—universal because, despite changes in the external environment, internal states consistently provide contextual information that guides adaptive behavior, and valuable because they are intrinsically linked to the reward system[11,12]. Furthermore, by incorporating insights from robotics [13-15], AI [16], and neuroscience [17-20], we emphasize the significance of modulatory computations that leverage the internal state as a context, suggesting a promising direction for advancing autonomous and adaptive AI. We argue that this framework holds the potential to address critical challenges in RL research, such as the exploration-exploitation tradeoff and the stability-plasticity dilemma, while also serving as a computational framework for interoception and affect [21,22].

## Box 1. Theoretical Studies of Life

"What is life?" is a long-standing question [23], and to address this, a systems view of life has been proposed [24]. The systems view of life considers a living organism as an integrated whole rather than the sum of its parts. Ludwig von Bertalanffy, a pioneer of general systems theory, pointed out that every living organism is an open system, far from equilibrium, resisting the dissipative nature of physics (i.e., the second law of thermodynamics) [25]. An open system exchanges energy and matter with its surroundings, and a self-organizing living system dynamically interacts with its environment. Following this view, Ross Ashby tried to abstract the essentials of the "machine in general" from the physiological and metabolic concept of living organisms [11]. He suggested that materiality needs to be removed from defining the core concept of machines in general. Excluding materiality, he defined the concept of survival in a mathematical form based on essential variables, which should be kept within bounds. According to his definition of survival, the key problem becomes how a system (i.e., general machine) can maintain its essential variables within bounds when faced with external perturbations. Then, survival becomes "the problem of adaptation" [26]; when external perturbations impact a system by moving its essential variables away from the set point (or more generally, attracting set), the system needs to take adaptive actions to bring the essential variables back to the set point (i.e., homeostasis). Maturana and Varela suggested the concept of autopoiesis to extend Ashby's views to encompass the phenomenon of life [26,27]. Their key interest was the emergence of biological autonomy [28,29]. Instead of the stability of living organisms, they focused on the organisms' active control for survival, and they recognized this as the central question of biology [26]. Recent research on life spans diverse fields, including astrobiology [30,31], which investigates the origins and conditions for life across the universe, and artificial life [32,33], which seeks to understand life by recreating biological phenomena through simulations and robotics. These advances provide perspectives consistent with our interoceptive AI framework, highlighting the significance of interactions between systems and their surroundings, as well as the balance between autonomy and environmental engagement[29].

The free energy principle (FEP) is another recent theoretical framework for life and cognition [34] and describes the characteristics of life in a principled way. The FEP emphasizes the importance of separating an organism's internal state from its surroundings, offering a formal framework to define the separation using the concept of conditional independence [12]. Conditional independence implies that, given knowledge of the boundary state ($s^B$), the system's internal state ($s^I$) is conditionally independent of its external state ($s^E$): ($s^I \perp s^E)|s^B \iff p(s) = p(s^I|s^B)p(s^E|s^B)p(s^B)$. This relationship underscores the mediating role of the boundary in decoupling internal and external states. Furthermore, the FEP proposes that all living organisms minimize the surprisal (a.k.a., self-information) of their exchanges with the external environment. However, the average surprisal corresponds to entropy, an information-theoretic quantity that cannot be evaluated directly. Thus, the FEP suggests that organisms minimize an upper bound on surprisal, known as variational free energy, thereby resisting increases in entropy. The FEP formalizes an organism as a random dynamical system and offers a normative account of self-organization from a Bayesian perspective. This perspective rests upon the ability to separate internal states from external ones through the maintenance of a boundary, known as a **Markov blanket** [12], which is characterized by sparse coupling between internal and external states. Self-organizing biological systems that maintain their boundary effectively minimize their entropy. Although the FEP is grounded in a rigorous mathematical framework [12], which extends beyond the scope of this article, its core idea can be recapitulated into "control-oriented predictive regulation." This can be derived from the basic principles of physics and potentially applicable to all living systems that maintain the integrity of their Markov blanket by avoiding surprising exchanges with the external environment. The FEP has been applied to various aspects of self-organization and sentience [35-37], particularly through its operationalization in the form of active inference.

## INTEROCEPTION FOR NATURAL AND ARTIFICIAL AGENTS

Interoception indicates a biological process to monitor the physiological variables constituting the internal states (or internal milieu [38]), such as glucose levels, oxygen levels, and blood pressure [39]. This internally-oriented process is, as the name indicates, a contrasting concept to its counterpart, "exteroception", a typically-known perceptual process to monitor sensory stimuli in the external environment. In neuroscience, interoception has been actively studied in conjunction with other mental faculties, such as feelings [40-42], emotions [43-45], cognition [46], psychopathology [47], and consciousness [48], highlighting its pivotal roles across various mental functions of agents (**Fig. 1B**). More recently, abstracted as a set of processes that allow a given system to "sense, interpret, integrate, and regulate signals from within itself," [49] the core functionality of interoception has been adapted in diverse engineering fields. For instance, in the field of space robotics and aerospace, the methods for monitoring the structural integrity of the hardware, such as plastic deformations and fatigue damage, or overall system fault detections, have been studied and implemented, yet under different names, such as "Prognostics and Health Management" or "Integrated System Health Management" [50,51]. Modern AI technologies, such as deep learning, have recently been integrated with these systems to predict and manage complex system health [52-54].

A critical aspect of interoception lies in its integrative and regulatory functionality [17,55,56]. For example, natural agents (i.e., living organisms) continuously monitor their internal physiological states while performing complex tasks and adapt their actions based on their internal states. Significant deviations in variables such as glucose levels or body temperature prompt animals to prioritize restoring these variables to homeostatic ranges. This results in autonomous shifts in goals, such as switching from exploring the environment to foraging for food or seeking a suitable place to regulate body temperature. This process is adaptive, as animals do not overwrite existing knowledge to switch their goals or behaviors. Instead, they hierarchically modulate their biological neural networks to select appropriate actions or goals to address pressing issues. The neuroanatomical structure of the interoceptive signal pathway, which passes through the brain's modulatory centers, is positioned to influence higher-order association brain areas [17], including the prefrontal cortex, insula, and anterior cingulate cortex, regions important for abstracting value representations [57] and multimodal integration [58]. These characteristics may provide valuable insights for the development of autonomous and adaptive artificial agents.

Previous research in robotics has introduced artificial neuromodulatory and neuroendocrine systems to enhance autonomy and adaptivity, yet without fully leveraging interoceptive mechanisms in living organisms [59-62]. For instance, it has been demonstrated that neuromodulatory systems can be introduced to induce behavioral changes in robots, allowing them to adapt more effectively to their environment by dynamically modifying behavior in response to newly incoming information [63]. Similarly, artificial neuroendocrine systems have been employed to adjust parameters, emulating hormonal mechanisms to modulate a robot's sensitivity to stimuli [60]. These hormonal adjustments can regulate tolerances and sensitivities to both internal and external conditions (e.g., a robot in an energy-rich environment tolerates larger energy deficits), thereby promoting adaptation to diverse environments. Despite these advancements, however, most studies remain focused on specific tasks or algorithms, lacking an integrative framework that incorporates ideas from neurobiological mechanisms of interoception into the design of artificial agents or robots to foster the development of more autonomous and adaptive agents [8,9].

## BASIC ELEMENTS FOR IMPLEMENTING INTERNAL STATES

The first step for developing interoceptive AI is the implementation of internal states in artificial agents. This can be achieved by adopting a functional interpretation of internal states, drawing on concepts from cybernetics and modern theories of life. Such an approach requires a shift from a spatio-anatomical perspective (i.e., internal states confined within a biological body) to a functionalist view, acknowledging that natural and artificial agents can have distinct forms of internal states, with artificial agents operating at an abstract level. The basic functional characteristics of internal states include 1) factorization, 2) stable state dynamics, and 3) being a source of primary reward.

The first characteristic, "factorization" with respect to the external environment, is a fundamental feature of living organisms. For example, internal states in humans, such as body temperature, are 'factorized' or separated from external variables, e.g., ambient temperature. This factorization allows agents to insulate their internal variables effectively, thereby establishing a clear boundary between themselves and their surroundings. By explicitly considering internal states, the interoceptive AI framework endows an agent with an internal milieu that is distinct from its external environment. Specifically, the factorization enforces internal and external states to have their own state dynamics [64]—for instance, the dynamics of body temperature can be distinguished from those of environmental temperature. Importantly, the factorization does not eliminate the interactions between these states. For example, ambient temperature can influence body temperature, but this interaction is mediated by boundary states, such as the skin, allowing for selective influence of external states on internal ones, unlike the direct interactions among external states. This selective interaction contributes to the second characteristic of internal states: relatively stable compared to those of the external states. For instance, while the temperature of the external environment may vary throughout the year, the human body consistently maintains a temperature of approximately 36.5°C. However, this stability does not arise automatically from factorization; it requires active regulation through negative feedback mechanisms—a process known as homeostasis [65]. Furthermore, successful regulation requires active engagement with the external environment while seamlessly integrating internal and external information. Lastly, maintaining homeostasis serves as the primary source of reward for living organisms—i.e., survival. Deviations of internal states beyond their optimal range, known as the viability zone, lead to failure to survive. In this sense, internal states can be viewed as a set of variables that define the survival of an agent.

These characteristics are inspired by living organisms, laying the groundwork for understanding how internal state variables guide and regulate their internal and external actions in real-life situations. Internal states with these characteristics have the potential to serve as a universal and intrinsically valuable context, meaning that they provide consistent and structured information that is important for survival and remains relatively stable despite external variability. In the case of living organisms, internal physiological states such as glucose levels, body temperature, and other physiological states act as reference points for interpreting the environment and determining appropriate actions. By relying on the internal state information, living organisms effectively predict and calculate rewards and adapt to continuously changing external environments [55].

These ideas are not entirely new, with antecedents in earlier literature, but their further development and exploration have been limited, likely due to their origins in distinct and unconnected fields and the lack of an integrative framework. For example, the idea of establishing a functional and computational link between internal state variables and survival was first proposed in cybernetics by W. Ross Ashby [11]. He described an organism as a dynamical system and introduced the notion of essential variables—an abstracted form of internal state variables that define survival (**Box 1**). Examples of essential variables in animals include body temperature, glucose levels, and blood pressure, all of which must be maintained within specific bounds to ensure survival [66]. A related idea emerged from the RL field, where Singh et al. [67] proposed to divide state variables into internal and external ones, with internal states providing the reward signal. More recent studies have also explored the concept of homeostasis in the context of RL by incorporating the idea of internal states [65,68]. Building on and integrating these ideas, our interoceptive AI framework seeks to offer a more integrative and computational formalization of internal states and interoception and their implications.

We can achieve this by formalizing the internal states within the Markov Decision Process (MDP) framework, which then can be applied to diverse domains, such as transportation, manufacturing, and financial modeling, etc., providing a novel perspective on the functionality of the internal states beyond physically embodied agents (**Box 2**) [69]. The basic setup for implementing internal states in the MDP framework involves three major steps: 1) factorizing internal states by separating their state transition probabilities from those of external states, 2) applying selective (or sparse) interactions between internal and external state transition probabilities, and 3) mapping the primary reward function onto the dynamics of internal states (**Fig. 2**). These steps represent the bare minimum for the implementation of internal states, providing a foundational framework upon which researchers can incorporate additional interoception- and life-inspired features into MDPs. The aim of this paper is not to provide an exhaustive list of such features but rather to offer a basic framework with open implementation ideas, providing researchers with the flexibility to expand and adapt it to suit their specific objectives. In this sense, the versatility of the MDP framework makes it well-suited to this aim, as it can flexibly accommodate a variety of algorithmic settings by introducing any mathematical constraints. For example, homeostatic properties can be integrated into the reward function, as has already been developed as homeostatic RL [65]. This allows agents to regulate their internal states within the defined bounds through negative feedback, thereby ensuring more stable internal state dynamics. In addition, neuromodulatory functions can be implemented to introduce system-level regulation, which will be discussed in more detail in the next section.

### Box 2. Markov Decision Process

Markov Decision Processes (MDPs) are a well-established framework for formalizing sequential decision-making problems [70]. It is defined with a set of states $s \in S$ at any given time step, and at each step, an agent selects an action $a \in A$ to maximize the total reward (or minimize cost). In MDPs, the state transitions occur based on a state transition probability, which defines the environment's dynamics. This probability determines how likely a previous state $s$ transitions to the next state $s'$, given that the agent performs an action $a$: $p(s'|s,a) \doteq P\{S_{t+1} = s' \mid S_t = s, A_t = a\}$. The action results in a reward for an agent at each step. The reward is determined by a reward function that assigns a single numeric value, denoted as $r$, to each state-action pair: $r_t = R(s_t, a_t)$. The objective of an MDP is to select an action (or a sequence of actions) that maximizes the cumulative reward in the long run. Reinforcement learning (RL), a branch of AI research, relies on MDPs to solve various decision-making problems. Many RL algorithms utilize a value function, which represents the total amount of expected reward starting from the initial state $s$ while following a particular policy $\pi$ (the probability of selecting possible actions given a state): $v_\pi(s) \doteq E_\pi \{\sum_{k=0}^{\infty} \gamma^k R_{t+k+1} \mid S_t = s\}$, where $\gamma$ is a discount factor for future rewards. While the components of MDPs, such as states, state transition probabilities, and reward functions, can be flexibly configured to meet research objectives, it is essential to adhere to the mathematical constraints of MDPs. For example, the MDP framework is constrained by the Markov property, which posits that each state encapsulates all relevant historical information. In cases where states are not fully observable, the Partially Observable Markov Decision Process (POMDP) can be used [71]. POMDPs introduce an observation function (a.k.a. likelihood or emission function) to establish a probabilistic mapping between states and observations, enabling agents to infer states based on observations. Further expanding on POMDPs, the Mixed Observability Markov Decision Process (MOMDP) incorporates states with varying levels of observability [72]. This adaptability highlights the versatility of the MDP framework, allowing it to accommodate diverse problem settings by introducing different types of mathematical constraints.

Although conventional RL models typically assumes stationary MDPs, in which state transition probabilities and reward functions remain unchanged over time [73], this stationary assumption is often violated in real-world scenarios [7,74]. Living organisms, for instance, frequently encounter situations where their tasks and goals are continuously changing (i.e., non-stationary reward functions), and their environments are also changing (i.e., non-stationary state transition probabilities). To address this non-stationarity problem, multiple approaches have been proposed (for a comprehensive review on this topic, see [7]), with some using factorized task state variables to represent dynamically changing tasks [74,75]. Our interoceptive AI framework adopts a similar strategy by factorizing the state space into internal and external states, with internal states having greater stability. A unique aspect of our approach lies in the interpretation of stationary components as internal physiological states, in which we endow with multiple life-inspired features [12,65,68].

## NEUROMODULATION FOR INTEROCEPTIVE AI

Neuromodulatory functions serve as a key mechanism that bridges internal states with autonomous and adaptive behaviors, as previous neuroscience literature has highlighted the role of a neuromodulatory process in mediating the interactions between the body and the brain to maintain homeostasis and facilitate adaptive responses to dynamically changing environments [17,20,55]. In Drosophila, for example, several studies have shown that internal states such as hunger can modulate sensory processing through neuromodulatory mechanisms, influencing foraging behavior. During starvation, short neuropeptide F enhances sensitivity to food-related odors via the DM1 glomerulus, while tachykinin suppresses aversive responses in the DM5 glomerulus, enabling adaptive foraging [76]. In addition, beyond sensory-driven behaviors, internal states can dynamically shape the balance between exploration and exploitation, with nutrient-deprived Drosophila tending to exploit specific food patches while satiated flies engage in broader exploration, likely optimizing future foraging opportunities [77,78]. Moreover, metabolic cues influence memory formation, as hunger modulates both encoding and retrieval of learned food associations through neuromodulatory circuits in the mushroom bodies[79-81].

To implement modulatory mechanisms in artificial systems, it is essential to abstract their computational functionalities, which primarily involve modulating neuronal sensitivity to inputs [82]. These computations ensure the efficient encoding of sensory information through two key mechanisms: multiplicative gain modulation, which amplifies relevant signals to enhance signal-to-noise ratios, and additive gain modulation, which adjusts neural excitability to maintain a balance between responsiveness to new inputs and the stability of ongoing processes. Combined with interoceptive inputs—proposed as universal and valuable contexts—these mechanisms could play a crucial role in enabling context-dependent adaptive responses by dynamically regulating the system's functionalities [83].

The neuromodulatory mechanisms have been applied in robotics [62,84] and reinforcement learning [85,86] to design adaptive robots and AI agents. For instance, neuromodulators, such as dopamine, serotonin, acetylcholine, and noradrenaline, have often been formalized as hyperparameters, with neuromodulation conceptualized as the tuning of hyperparameters, including reward sensitivity, exploration-exploitation balance, and learning rates [62,85]. In addition, a Bayesian interpretation of neuromodulatory computation suggests that neuromodulators encode uncertainty or its complement, precision [20]. These concepts have recently been extended to deep neural networks, incorporating a primary network for task processing and an auxiliary neuromodulatory network that dynamically adjusts the primary network's weights and activations based on contextual inputs. This architecture has been shown to mitigate catastrophic forgetting by regulating synaptic plasticity, enabling the system to preserve previously learned knowledge while accommodating new information via the selection of sub-networks in a context-dependent manner [87-89].

Interoceptive AI incorporating modulatory mechanisms has the potential to address several longstanding dilemmas in AI. One such dilemma is the exploration-exploitation trade-off, in which agents risk missing potentially better outcomes by adhering to known actions (i.e., exploitation) or failing to utilize known actions effectively due to excessive exploration [70]. The common approach in RL to address this dilemma involves using ad hoc novelty-based intrinsic rewards, encouraging agents to explore more when encountering novel situations in their environment [90-93]. However, in non-stationary environments, this method can result in persistent exploration, as agents would continuously encounter novel situations [92,93].

The interoceptive AI framework offers a complementary approach, drawing inspiration from animal behavior literature, which shows that living agents decide whether to explore or exploit based on their internal needs [55,77]. For example, a hungry animal exploits its existing knowledge to locate food, whereas a satiated animal explores new behaviors to gain knowledge, effectively balancing exploration and exploitation based on its internal state. Several studies in both simulated and robotic systems have investigated mechanisms for achieving this balance[94-96]. While the specific implementation of such mechanisms within the deep neural networks remains an area for future investigation, modulatory mechanisms in this framework could dynamically adjust exploration and exploitation based on the system's internal state, similar to the way animals modulate their behavior through interoception-driven neuromodulation [77,97]. This approach offers a principled and context-sensitive alternative to ad hoc strategies.

In addition, the recently introduced energy landscape framework in neuroscience may offer a computational lens for understanding how modulating neural weights and activations could guide neural network transitions between exploitative and exploratory behaviors [97]. This approach formalizes the brain's states as positions in an energy landscape, where stable states correspond to valleys and less stable states to hills. According to this interpretation, the brain can resolve conflicting needs by navigating this landscape. For example, neural activity may shift from a region associated with hunger or thirst to another region representing exploratory behaviors as the organism's internal state evolves [97].

Another dilemma is the stability-plasticity trade-off, a long-standing challenge in machine learning [98] and RL research [7]. It arises when agents must balance retaining prior knowledge (stability) while incorporating new information (plasticity) in non-stationary environments. Current AI systems often suffer from catastrophic forgetting, where new learning overwrites previously acquired knowledge. In the conventional RL framework, agents operating in volatile environments struggle to determine which information to retain or update due to a lack of predefined assumptions regarding environmental stability.

The interoceptive AI framework has the potential to address this dilemma by leveraging the stability of internal states, even in the presence of a non-stationary external environment. For example, neuromodulation-based algorithms could implement context-dependent learning, allowing agents to select sub-networks in neural networks for different contextual cues [87]. Another recent study demonstrated the benefits of stable context representations in meta-learning scenarios, where agents must acquire new information while retaining prior knowledge [99]. By incorporating these ideas into the interoceptive AI framework, internal states could serve as a stable anchor for neuromodulatory algorithms, providing a robust foundation for adaptive and context-sensitive learning.

Inspiration can also be drawn from the brain's interoceptive systems, which naturally balance stability and flexibility. Evolutionarily conserved systems, such as the vagus nerve and brainstem, provide stable, reliable processing of interoceptive signals to regulate basic physiological states [100]. At the same time, higher-order regions like the insula, anterior cingulate cortex, and ventromedial prefrontal cortex enable flexible representations of values and goals, supporting adaptive responses to dynamic environments [57,58,101]. These neuroanatomical features of interoceptive brain systems can offer life-inspired strategies for designing adaptive AI systems capable of addressing the stability-plasticity dilemma by integrating stable and flexible processing mechanisms.

### INTEROCEPTIVE AI AS COMPUTATIONAL MODELS OF INTEROCEPTION AND AFFECT

The interoceptive AI framework could serve as a computational model for interoception and affect (**Fig. 3A**). With remarkable advances in AI, neuroscientists have begun to use AI agents as computational models of cognitive functions [102]. By comparing the AI agents' behaviors and neural representations with humans or animals, one can study the underlying neurocognitive principles that are often difficult to investigate using traditional experimental methods in neuroscience. However, implementing this approach requires well-defined tasks for AI. Compared to cognitive and decision-making processes, interoception and affect-related processes currently have limited tasks for testing. Recently, researchers have proposed that homeostasis and allostasis can offer key testable bases to study affective neuroscience [22]. Moreover, emotions and homeostasis have been studied in the context of robotics [13-15,103]. Building on these prior knowledge, interoceptive AI could serve as a new integrative framework that can be used to study key topics in affective neuroscience.

Previous studies have proposed computational models of interoception using RL and predictive processing (e.g., predictive coding and active inference) [21,104], both of which can be integrated into the interoceptive AI framework. Notably, active inference and interoceptive AI share many of the same commitments (**Fig. 3B**). While interoceptive AI focuses more on formalizing the internal states for their contextual roles, and active inference emphasizes the belief-updating or inference processes associated with self-organized behavior, both rest upon the constraints imposed by internal states or interoceptive modalities. The primary question addressed by active inference is, "How do living organisms persist while engaging in adaptive exchanges with their environment?" [34] Efforts to answer this question are in parallel to shaping the fundamental concepts of interoceptive AI. In addition, one of the unique contributions of active inference to the conventional RL framework is to replace the reward function with prior preferences [34], which, in the context of interoceptive AI, can be interpreted as internal set points or attracting sets of internal states. Active inference has also been compared with conventional RL approaches [36,105,106], suggesting its ability to effectively handle dynamically changing goals [107] and non-stationary external environments [37], again the two primary objectives that the interoceptive AI framework pursues. Recently, active inference has been applied to interoception, leading to the development of interoceptive active inference [45,108], which provides a computational account of interoception and affect [21,22,109].

Expanding on this foundation, precision weighting emerges as a key mechanism in active inference that is particularly relevant to interoceptive AI, as it governs the extent to which prediction errors influence belief updating [20]. In active inference, prediction errors are signals that indicate discrepancies between expected and actual sensory inputs. However, not all prediction errors are equally reliable; some carry more meaningful information than others[110]. This selective emphasis on certain prediction errors over others allows for more flexible and adaptive behavior, as the brain can prioritize important signals while filtering out noise. This mechanism is intricately linked to neuromodulation like dopamine, acetylcholine, noradrenaline, and serotonin, which regulate the precision of sensory signals, including interoception [20]. In active inference, precision modulation serves as a principled mechanism that integrates two distinct computational mechanisms—neural gain control and hyperparameter tuning. Neural gain control dynamically regulates the excitability of neural circuits, determining how strongly prediction errors influence belief updates. Simultaneously, precision modulation functions are akin to adjusting the generative model's parameters, allowing the system to adapt flexibly to non-stationary environments, much like hyperparameter tuning in RL. By incorporating these principles from active inference, the interoceptive AI framework seeks to provide a formalized model of interoceptive processes and affective states, advancing our understanding of their computational and neural underpinnings.

## CONCLUDING REMARKS

Drawing inspiration from living organisms, here we introduced the interoceptive AI framework, crafted to enhance the autonomy and adaptivity of artificial agents through the incorporation of an internal environment into the traditional AI framework. The proposed system enables an agent to monitor its internal state and adaptively recalibrate its goals and responses to cope with environmental changes. According to Claude Bernard, "the stability of the internal environment is the condition for the free and independent life," [111] and Singh et al. noted that "all rewards are internal." [112] We believe that our life-inspired ideas of state factorization, mapping rewards onto the internal state dynamics, and integrating neuromodulatory mechanisms will serve as essential building blocks for building autonomous and adaptive intelligence. More importantly, we aspire that our interoceptive AI framework will deepen our understanding of animal and human intelligence.

Notably, the capabilities enabled by interoception in AI systems could potentially raise ethical and moral considerations. Recognizing these concerns, we emphasize the importance of fostering careful discussions and building a consensus aligned with human values on ethical and governance issues with the development and application of interoceptive AI systems. Such efforts are essential to ensure that these technologies are developed and utilized in ways that uphold societal values and benefit humanity.

## ACKNOWLEDGMENTS

This work was supported by IBS-R015-D2 (Institute for Basic Science; to C.-W.W. and S.J.H.) and by HI19C1328, which is a grant of the Korea Health Technology R&D Project through the Korea Health Industry Development Institute (KHIDI), funded by the Ministry of Health &

Welfare (to S.W.L.)

**A** Challenges in training artificial agents

Dynamically changing environments (non-stationarity)

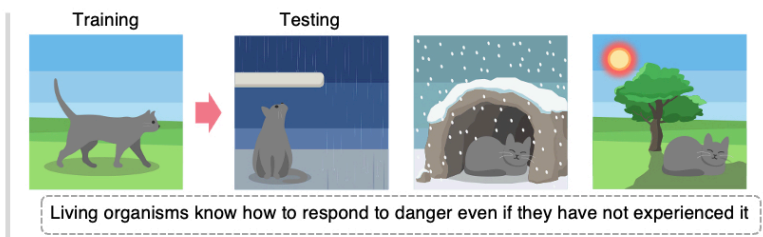

Living organisms know how to respond to danger even if they have not experienced it

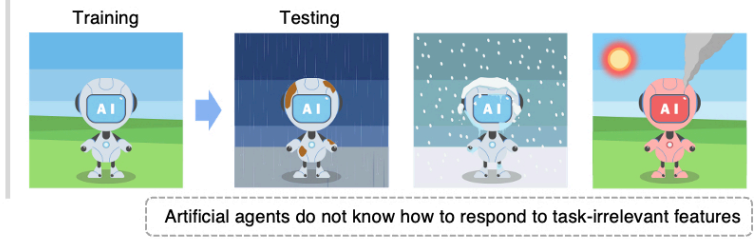

Artificial agents do not know how to respond to task-irrelevant features

Dynamically changing goals (context-dependent values)

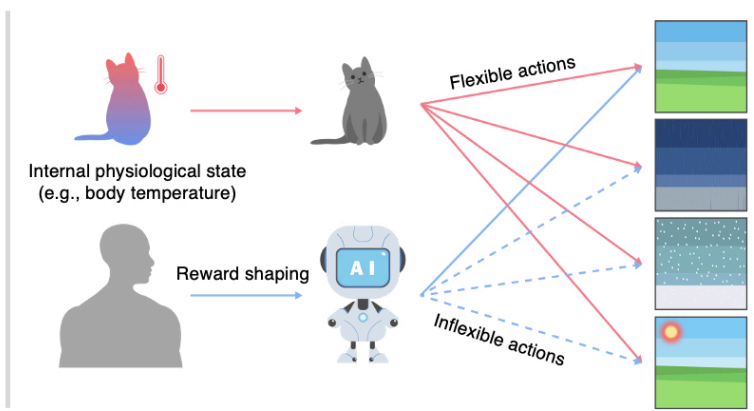

**B** Interoception in the brain

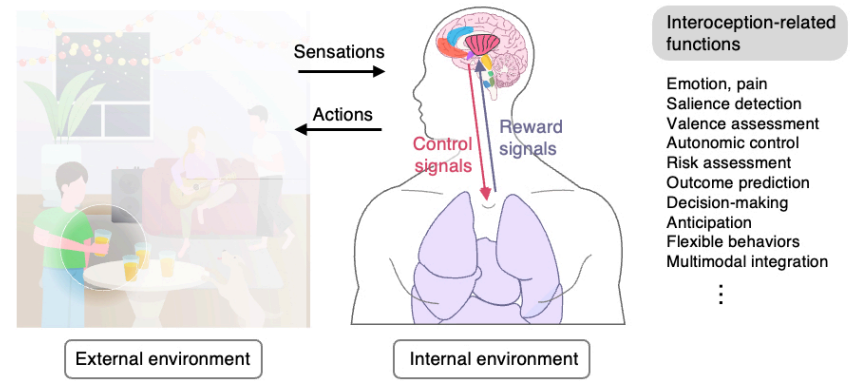

**C** Relevant reinforcement learning (RL) frameworks

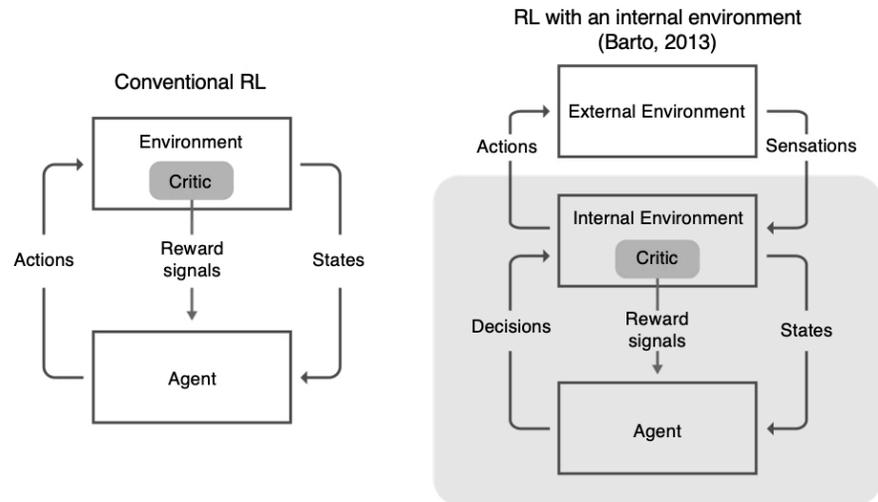

**Figure 1. Interoception for autonomous and adaptive agents. (A)** Examples of dynamically changing environments (top) and goals (bottom), which are the two main targets of the interoceptive AI framework. These are significant challenges in current AI systems. (top) Adaptivity: Animals can adapt to novel and changing environments by utilizing stable internal values (e.g., the cost of being too

hot, cold, or wet) without explicit training. Unlike animals, however, robots have no internal values and thus find it difficult to adapt to changing and new environments. (bottom) Autonomy: Animals can easily perform flexible actions with dynamically changing goals, but robots need a designer's additional instructions to change their goals. **(B)** Interoception plays a critical role in providing a stable and context-dependent reference for humans and animals. For example, reward signals are generated from the internal environment (purple arrow), which report an organism's homeostasis (e.g., thirst). Interoception then serves to contextualize exteroception in the brain, making some stimuli (e.g., beverages) more salient than others to achieve homeostasis (e.g., amount of water in the body). The brain can also send a descending control and regulatory signal (red arrow) to the body (e.g., allostasis) [113,114]. Interoception is known to play an essential role in diverse cognitive and affective functions and their interactions. Brain regions known to be important for interoception, such as the insula, medial prefrontal cortex, hypothalamus, and brainstem, are highlighted with different colors. (C) (left) The conventional reinforcement learning (RL) framework without an internal environment state. In the conventional RL, rewards stem from the external environment state. (right) A previous study [115] proposed an alternative model to study intrinsically motivated behaviors, which suggests that reward signals should come from the internal environment.

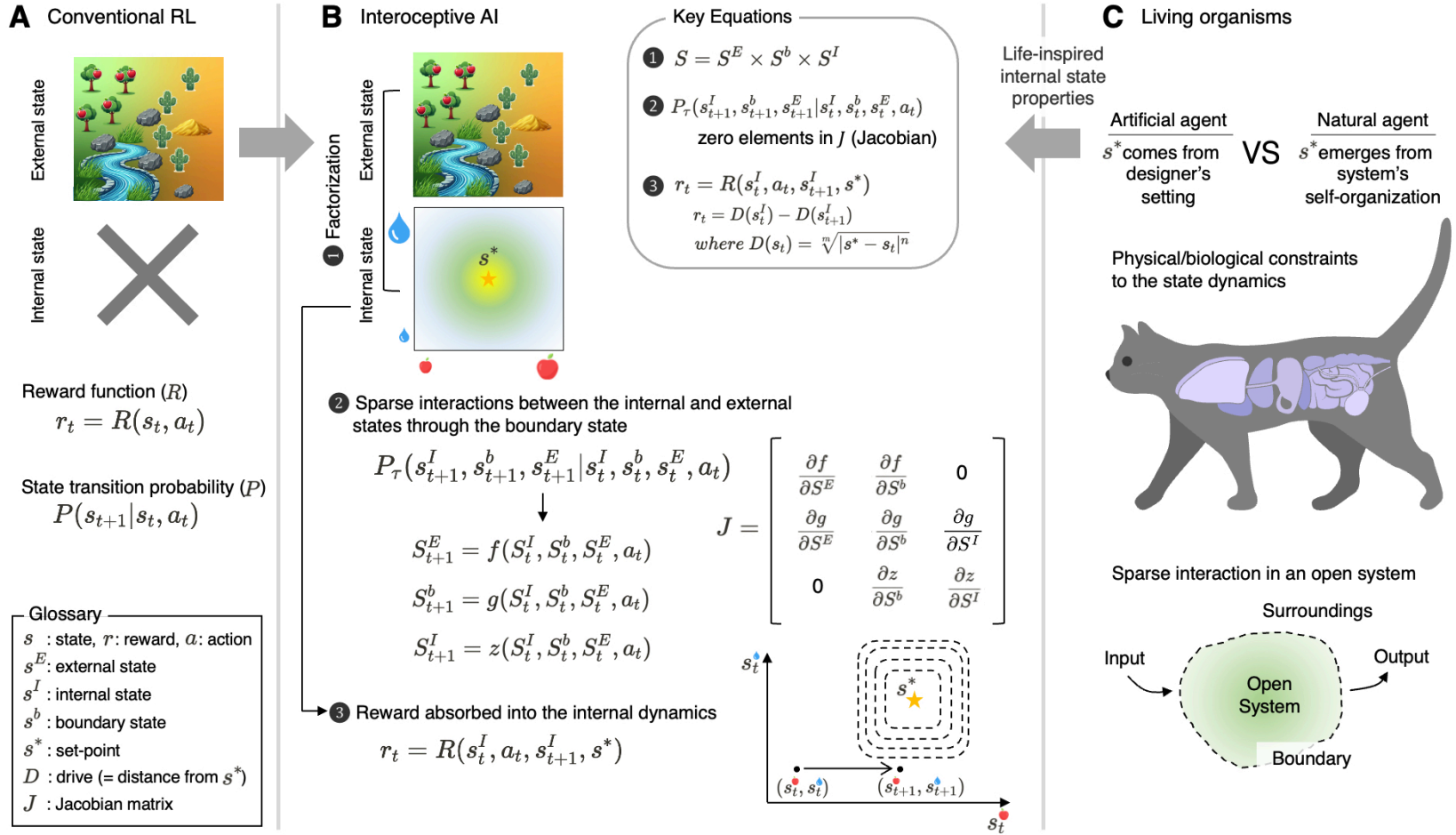

**Figure 2. Basic elements for implementing internal states.** This provides an example of interoceptive AI in MDPs. **(A)** Conventional RL models only consider external states. For example, Sutton and Barto's RL textbook says that "anything that cannot be changed arbitrarily by the agent is considered to be outside of it." [70] In this framework, the reward function is mapped onto the external states, while the state transition probability governs the dynamics of external states. **(B)** The basic setup for implementing internal states can be summarized as the following: 1) Interoceptive AI introduces the internal states into the model through state

factorization. The internal and external state dynamics are characterized by factorized state transition probabilities. 2) Interoceptive AI formalizes stability by structuring internal states to interact selectively or sparsely with external states through boundary states. This selective interaction is represented in a Jacobian matrix, with zero elements indicating the decoupling between internal and external variables given boundary states. 3) The reward originates from the internal states: reward is absorbed into the internal state dynamics because the reward can be expressed in terms of some measure of distance from an internal set point (i.e., attracting sets of internal states). **(C)** In living organisms, the set point (or set points) of the internal states underwrites self-organization and emerges through natural learning and selection [116]. In contrast, interoceptive AI remains somewhat artificial because its set point can be determined by a designer.

**A** Interoceptive AI as a computational model of interoception and affect

**B** Interoceptive active inference

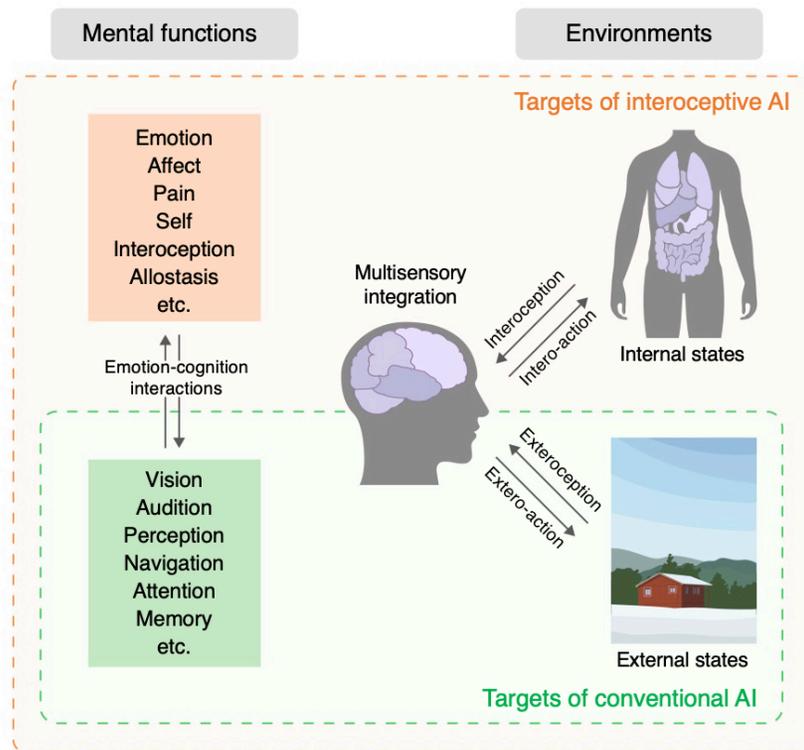

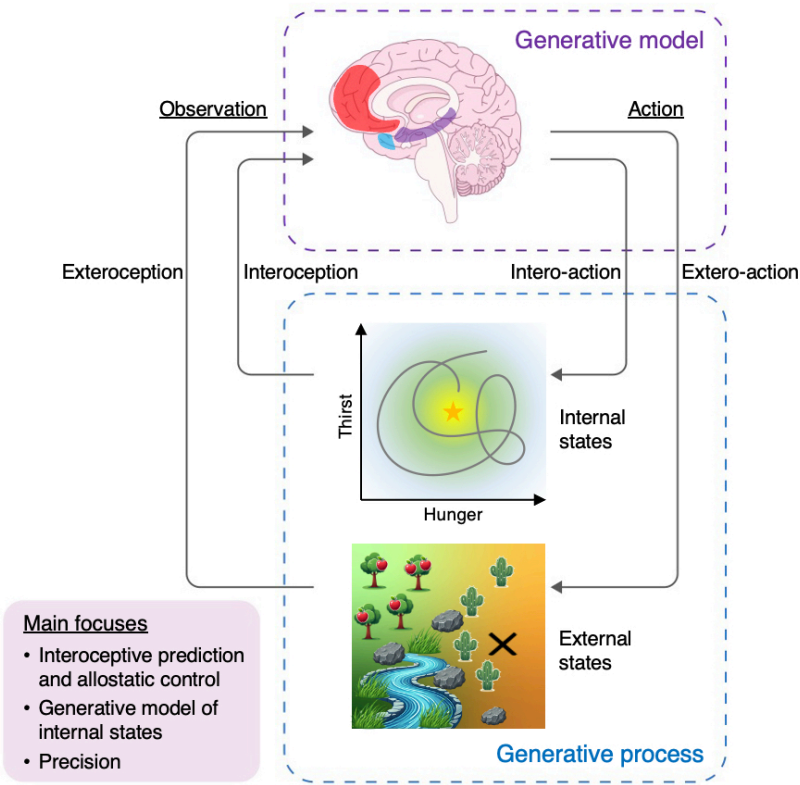

**Figure 3. The interoceptive AI framework for affective neuroscience. (A)** Conventional AI models have provided a useful framework for the study of perceptual and cognitive functions, such as vision, navigation, and memory (the green area). They usually consider only the mental functions and processes pertaining to extra-personal environments (e.g., exteroception), and thus they are not suitable for studying internal environment-related functions, such as emotion, pain, and interoception (the orange area). Alternatively, the interoceptive AI framework encompasses an internal environment, providing a computational model for interoception-related functions and their interactions with exteroception-related functions, such as value-based decision-making (the arrows between the orange and green boxes). **(B)** The interoceptive active inference is a computational model of interoception. According to interoceptive

active inference, an internal environment state is considered hidden from the brain. The brain only has access to the observations that the internal and external states generate (i.e., generative process), and based on the observations, the brain forms an internal model of how the observations are generated (i.e., generative model). In interoceptive active inference, the brain is assumed to engage autonomic reflexes to realize homeostatic set points for attracting sets. The distance between interoceptive inputs and set points is measured by the surprisal (or prediction error) in the same way that exteroceptive predictive coding uses surprisal or prediction errors for perception and proprioceptive prediction errors for motor function. The main focuses of Interoceptive active inference include belief updating required to generate appropriate (interoceptive) set points in the form of predictions and allostatic control and contextualizing the precision of accompanying prediction errors. Brain regions known to be important for internal generative models, such as the medial prefrontal cortex, entorhinal cortex, and hippocampus, are highlighted with different colors.